\documentclass[sigconf]{acmart} % for final version

\usepackage{multirow, makecell}			% For fancy table cell and multi-row tables
\usepackage{booktabs}					% For convenient booktabs
\usepackage{tabularx}					% For making fancy tables (IMPORTANT)
\usepackage{threeparttable}				% For footnote in table
\usepackage{siunitx}
\AtBeginDocument{%
  \providecommand\BibTeX{{%
    \normalfont B\kern-0.5em{\scshape i\kern-0.25em b}\kern-0.8em\TeX}}}

\copyrightyear{2021}
\acmYear{2021}
\setcopyright{acmcopyright}\acmConference[MM '21]{Proceedings of the 29th ACM
International Conference on Multimedia}{October 20--24, 2021}{Virtual Event, China}
\acmBooktitle{Proceedings of the 29th ACM International Conference on Multimedia
(MM '21), October 20--24, 2021, Virtual Event, China}
\acmPrice{15.00}
\acmDOI{10.1145/3474085.3475314}
\acmISBN{978-1-4503-8651-7/21/10}

\begin{document}
\fancyhead{}

\title{From Voxel to Point: IoU-guided 3D Object Detection for Point Cloud with Voxel-to-Point Decoder}

\author{Jiale Li}
\email{jialeli@zju.edu.cn}
\affiliation{%
%  \institution{College of Information Science and Electronic Engineering, Zhejiang University}
 \institution{Zhejiang University} 
 \city{Hangzhou}
 \country{China}}

\author{Hang Dai}
\authornote{Corresponding authors.}
\email{hang.dai@mbzuai.ac.ae}
\affiliation{%
  \institution{Mohamed bin Zayed University of Artificial Intelligence}
  % \streetaddress{1 Th{\o}rv{\"a}ld Circle}
  \city{Abu Dhabi}
  \country{United Arab Emirates}}

\author{Ling Shao}
\email{ling.shao@ieee.org}
\affiliation{%
  \institution{Inception Institute of Artificial Intelligence}
  \city{Abu Dhabi}
  \country{United Arab Emirates}
}

\author{Yong Ding}
\authornotemark[1]
\email{dingy@vlsi.zju.edu.cn}
\affiliation{%
%  \institution{College of Information Science and Electronic Engineering, Zhejiang University}
 \institution{Zhejiang University} 
 \city{Hangzhou}
 \country{China}}

\renewcommand{\shortauthors}{Li and Dai, et al.}

\begin{abstract}
In this paper, we present an Intersection-over-Union (IoU) guided two-stage 3D object detector with a voxel-to-point decoder. To preserve the necessary information from all raw points and maintain the high box recall in voxel based Region Proposal Network (RPN), we propose a residual voxel-to-point decoder to extract the point features in addition to the map-view features from the voxel based RPN. We use a 3D Region of Interest (RoI) alignment to crop and align the features with the proposal boxes for accurately perceiving the object position. The RoI-Aligned features are finally aggregated with the corner geometry embeddings that can provide the potentially missing corner information in the box refinement stage. We propose a simple and efficient method to align the estimated IoUs to the refined proposal boxes as a more relevant localization confidence. The comprehensive experiments on KITTI and Waymo Open Dataset demonstrate that our method achieves significant improvements with novel architectures against the existing methods. The code is available on Github URL\footnote{\url{https://github.com/jialeli1/From-Voxel-to-Point}}. 
\end{abstract}




\maketitle

\section{Introduction}
% This researching field is being rapidly promoted with the demands of practical tasks such as virtual reality, robotics and autonomous driving. 
% The Region Proposal Network (RPN) built on the map-view features achieves higher box recall than that built on the point features \cite{PointRCNN,STD}, which has been clearly stated in \cite{PVRCNN}. 
%The Region Proposal Network (RPN) built on the map-view features achieves higher box recall than that built on the point features \cite{PVRCNN}. 
Three dimensional object detector localizes objects with tight 3D bounding boxes. Compared with monocular and stereo image based approaches \cite{mono_3dod_acmmm19, luo2020c4av, dai2020commands,stereo_rcnn, luo2021m3dssd}, LiDAR based methods are more robust in autonomous driving \cite{KITTIDataset,waymo_dataset}. Current 3D object detectors mainly represent the point cloud as raw points or voxels. Point-based methods \cite{PointRCNN, 3DSSD} argue that the raw points preserve the pose information of objects, which is crucial to accurate localization. Voxel-based methods \cite{Second,SASSD} voxelize the point cloud in a voxel encoder and project the voxel features onto Bird's Eye View (BEV) as the map-view features. The Region Proposal Network (RPN) built on the map-view features \cite{PVRCNN, PartA2_TPAMI} can achieve higher box recall than that built on the point features \cite{PointRCNN, STD}. 

%Since the structure information of the objects are abstracted to the map-view features with the downsampling in the voxel encoder \cite{Second}, the voxel based methods suffer from information loss  \cite{PVRCNN}. To avoid this and maintain the high box recall in the RPN stage, we propose a novel voxel-to-point decoder after the voxel encoder to provide the 3D point features. 
The structure information of the objects are abstracted to the map-view features with downsampling in the voxel encoder \cite{Second}, which results in the lose of the detailed pose information in the raw points. To avoid this and maintain the high box recall in the RPN stage, we propose a novel voxel-to-point decoder to extract the discriminative 3D point features from the raw points. The voxel-to-point decoder consists of stacked residual voxel-to-point decoding blocks with skip connections \cite{Unet}. In each decoding block, we aggregate the features of the voxels that are surrounding the raw point as the voxel-to-point features. In the stacked decoding blocks, We use a residual learning \cite{ResNet} between the point features from the previous level and the current voxel-to-point features to extract enhanced point features that gradually become fine-grained through the hierarchical feature aggregation. A segmentation mask is used to supervise the point feature learning, which makes it sensitive to the foreground and the background information. We propose a 3D Region of Interest (RoI) alignment to align the point features and the map-view features with the proposal boxes for accurately perceiving the object position. Since the eight corners of the 3D bounding boxes do not always exist in the raw points, the RoI-Aligned features are finally aggregated with the corner geometry embeddings that provide the missing corner information. %in the detection refinement stage.

% Why not use a point-based network to provide the point features from the raw points? In our method, each voxel has grouped the internal points together in the voxel encoder for enlarging the receptive field \cite{VoxelNet}. Further, we aggregate the features of the voxels that are in the $K$-nearest neighbors of the raw point as the voxel-to-point features. The residual learning enables the extraction of more informative point features than those directly from raw points using point based networks \cite{PointNet, PointNet++}. The voxel-to-point decoder is more efficient that the decoder can directly use features from the voxel encoder instead of one more point-based encoder for encoding the raw points.

% Why not use another point-based network like \cite{PointNet++, PointNet, l2g_seg_pc_acmmm19} to provide the point features starting from the raw points? 
The reasons why not use another point-based network like \cite{PointNet++, PointNet, l2g_seg_pc_acmmm19} to provide the point features starting from the raw points are as follows. In our method, we aggregate the features of the voxels that are in the $K$-nearest neighbors of the raw point as the voxel-to-point features. Each voxel has grouped the internal points together in the voxel encoder \cite{VoxelNet} for enlarging the receptive field \cite{PointNet++}, which is more informative than the raw point aggregation. The voxel encoder is shared in both the map-view feature learning and the point feature learning, which is more computationally efficient.

% points 在voxel 里 group， more informative. With the voxel-to-point residual path. 基于more informative voxel feature的points decoding. \cite{Maskrcnn} 

% 都是用localization accuracy 和 localization confidence
In a two-stage object detector, the second stage takes the proposal boxes from the RPN stage as input to classify the foreground object proposals and predict the residuals to their ground truth for further box refinement \cite{fasterrcnn}. The classification scores are typically used as the metric to rank the proposal boxes for removing the redundant boxes in the Non-Maximum Suppression (NMS) procedure. Since classification and localization are solved separately, the localization confidence is always absent in the object detection pipeline. The IoU is a natural criterion for localization accuracy as the ranking criterion in NMS. Inspired by \cite{IoUNet,STD}, we train a 3D IoU estimation branch that is parallel to other regression branches. However, there exists a misalignment between the estimated IoUs and the refined boxes in the existing methods. Since the estimated IoUs are regressed from RoI-Aligned features using the proposal boxes, the estimated IoUs are aligned to the proposal boxes, not the refined proposal boxes. Thus, the estimated IoUs from the proposal boxes cannot be used as the localization confidence for the refined proposal boxes in NMS for box de-redundancy. %Such a misalignment prevent us employing the unaligned estimated IoUs as the detection confidence to conduct NMS with detection performance improvements.

A straightforward approach is to train the IoU estimation branch \cite{IoUNet} using the actual IoUs of the refined boxes. However, there are two issues: i) Since the refined boxes are always one stage later than the proposal boxes, we cannot obtain the refined boxes in the training of the refinement stage. We need an additional stage to train a new IoU estimation branch for the refined boxes, which brings more network parameters and computational costs. ii) Although we can infer the box regression branch to refine the proposal boxes and compute the actual IoUs of the refined boxes for training a parallel IoU estimation branch, there raises another misalignment between the RoI-Aligned features and the IoU training labels. The IoUs of the refined box should be regressed from the RoI-Aligned features of the refined boxes other than that of the proposal boxes. With a well-trained parallel IoU estimation branch, we can resolve this misalignment simply and efficiently by updating the proposal boxes with the refined proposal boxes with one more inference stage. The second-time estimated IoUs are aligned to the refined proposal boxes and used as the localization confidence. %to suppress the redundant refined boxes in NMS.

\section{Related Work}
\textbf{Grid-based 3D Object Detection.}
Grid-based methods convert the point clouds of unstructured data format to the regular grids like pixels and voxels for 2D or 3D convolutional processing. The early work MV3D \cite{MV3D} projects the points as pixels in an image of bird's-eye view (BEV) for feature extraction and 3D bounding box proposal with efficient 2D CNN. The following works \cite{AVOD,Contifuse} leverage the camera image features to compensate the BEV point cloud features with effective fusion strategies. Besides, VoxelNet \cite{VoxelNet} divides the points into small 3D voxels for 3D CNN and PointPillars \cite{PointPillars} constructs the pseudo-images after voxelization. Due to the high computational cost caused by a large number of empty voxels, 3D sparse CNN \cite{sparseconv,submanifold} is introduced for efficient computation by SECOND \cite{Second}. Based on \cite{Second}, PartA2 \cite{PartA2_TPAMI} explores the object part locations for finer 3D structure information learning. Discretizing the points to grids with limited resolution brings computational efficiency, but it weakens the information interpretation.

\begin{figure*}[!htp]%figure* for wider figures.
	\centering
	\includegraphics[width=0.95\linewidth]{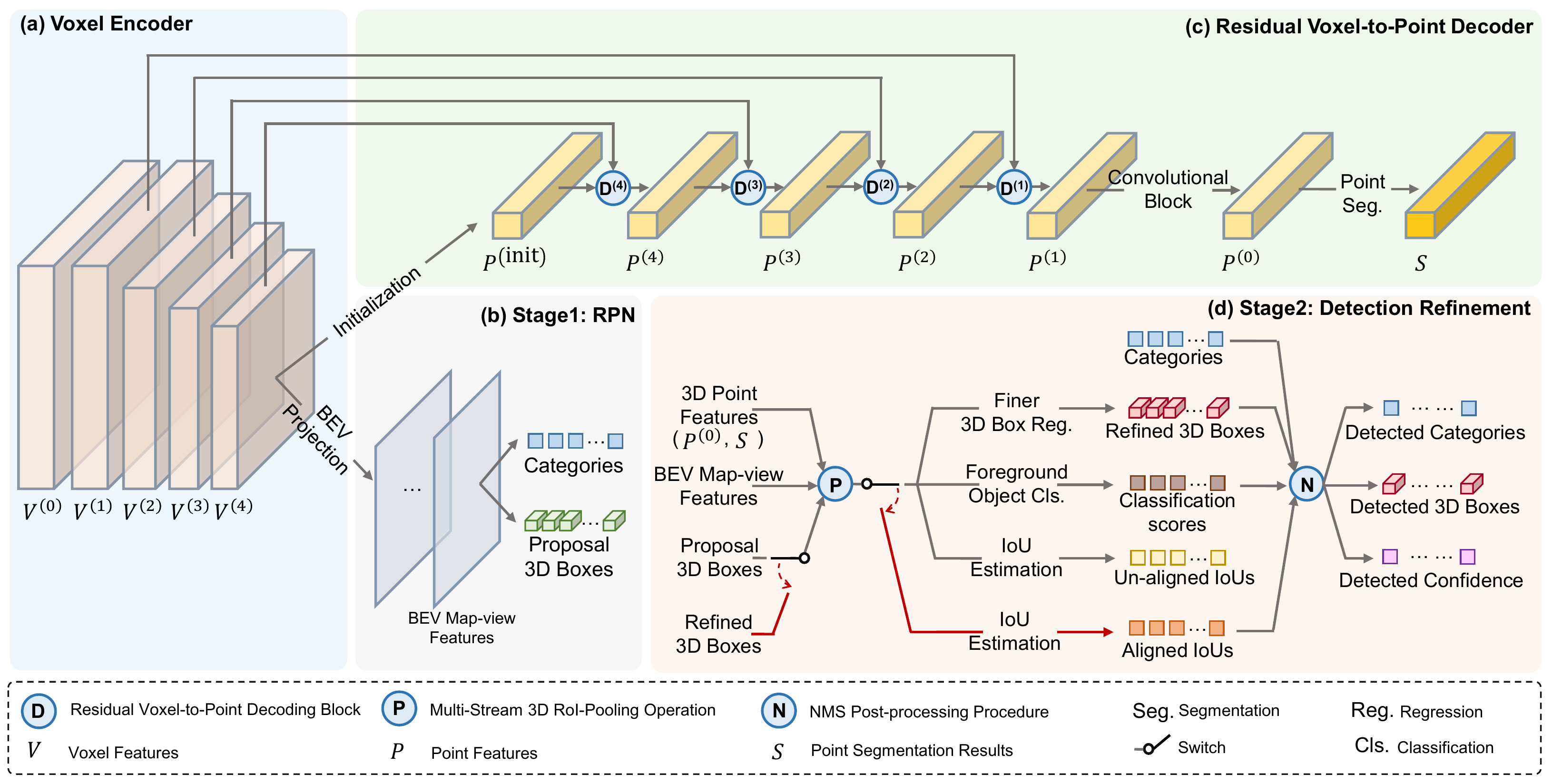}% Fig2_overallarch_updated
\vspace{-3mm}
	\caption{The framework of the proposed 3D object detection network. The red arrows represent the second-time inference by switching from the proposal boxes to the refined boxes for the 3D IoU alignment. The figure is best viewed in color.}
	\label{Framework}
\vspace{-3mm}
\end{figure*}

\textbf{Point-based 3D Object Detection.}
Point based methods take the raw point cloud as input, and apply PointNet++ \cite{PointNet++} or Graph Neural Networks (GNN) \cite{GNN_survey} for point-wise feature learning and object detection. PointNet++ based methods \cite{PointRCNN,li20203d,STD} maintain the resolution of the point cloud, while 3DSSD \cite{3DSSD} downsamples a relatively balanced number of foreground and background points, and discards the upsampling process. Different from the hierarchical feature aggregation, Point-GNN \cite{PointGNN} utilize GNN to iteratively update the features of the entire point cloud. Precise point coordinates are directly embedded into features, but it is not efficient in point sampling and grouping with a low box recall. 

\textbf{Voxel-Point based 3D Object Detection.}
The voxel-point based methods \citep{PVRCNN, STD, li2021p2v} use both representations. PointsPool \cite{STD} voxelizes the point cloud around the object proposal to encode the empty and non-empty regions for compact proposal-wise feature learning. PV-RCNN \cite{PVRCNN} proposes a Voxel Set Abstraction operation to aggregate the voxel-wise features in the backbone to some sampled keypoints. Then the keypoints are utilized for the bounding box refinement. Similarly, SA-SSD \cite{SASSD} converts the voxel-wise backbone features back to the point-wise features only for auxiliary supervision, which is not an effective way to exploit the point-wise features with accurate position information.

% we maintain the full point features of the raw point in an efficient way by searching the surrounding voxels for the raw point and aggregating the features of these voxels, then refine them in a residual manner with significant improvement, especially for the distant objects with much more sparse points.
\textbf{Ours VS. PV-RCNN.} In PV-RCNN \cite{PVRCNN}, a set of keypoints (2,048 points) is sampled to roughly represent the entire point cloud scene ($\sim$20K points), while only a small number of sampled keypoints distributed near the foreground objects can be used for the box refinement. To preserve more object details, we maintain the full point features of the raw points with our voxel-to-point decoder for the detection refinement with significant improvement, especially for the distant objects with much more sparse points. In PV-RCNN, voxel features of different levels are simply concatenated to each sampled point, while our decoder gradually enhances the point features through the hierarchical feature aggregation with residual learning. Unlike PV-RCNN that aligns the map-view feature to the several sampled keypoints, we align the map-view features to the evenly distributed grid points inside the object to focus more on the object region as shown in Figure~\ref{multi_stream_roi_align} (b). Besides, the absolute coordinates are used in the pooling operation of PV-RCNN, which is not robust to rigid transformations. Instead, we normalize both the coordinates of the raw points and grid points as the relative coordinates that are centered on the object proposal and aligned to the object proposal's orientation. %The relative coordinates are generally more robust. 

\section{Method} %Methodology
This section presents our IoU-guided 3D object detection with voxel-to-point decoder. We first explicitly define the point cloud 3D object detection task in Section~\ref{sec:problem_def}. Then the overall network structure is introduced in Section~\ref{sec:overall_arch}. The voxel encoder and RPN are presented in Section\ref{sec:vfe_rpn}. In Section~\ref{sec:res_v2p} and~\ref{sec:iou_guided_refine}, we describe a residual voxel-to-point decoder and an IoU-guided detection refinement in detail. The loss functions are presented in Section~\ref{sec:loss_func}.

\subsection{Problem Definition}\label{sec:problem_def}
The network takes one frame of the point cloud as input to localize the objects in the form of 3D boxes with localization confidence. The 3D box is represented as the 3D center point $(o_{{x}}, o_{{y}}, o_{{z}})$, size $(l, w, h)$, and rotation angle $\theta_{\text{rot}}$ around $Z$-axis. The localization confidence between 0 and 1 is also estimated by the network. Let $\left\{p_{i}=\left(x_{i}, y_{i}, z_{i}\right): i=1, \ldots, N\right\}$ be the coordinates of point cloud in the range of $\left[ {X}_{\text{min}}, {{X}}_{\text{max}} \right], \left[ {{Y}}_{\text{min}}, {{Y}}_{\text{max}} \right], \left[ {{Z}}_{\text{min}}, {{Z}}_{\text{max}} \right]$.

\subsection{Overall Network Architecture}\label{sec:overall_arch}
Figure~\ref{Framework} describes the overall network with four components: (a) the voxel encoder that extracts 3D voxel features; (b) an RPN where 3D voxel features are projected onto BEV as the map-view features for the object classification, and the localization of the coarse 3D box as the proposal box; (c) the voxel-to-point decoder that extracts the fine-grained point features from the voxel features; (d) the IoU-guided detection refinement stage for refining the proposal boxes and estimating the localization confidence. 

\subsection{Voxel Encoder and RPN}\label{sec:vfe_rpn}
The point cloud is voxelized by a quantization step $d=[d_{x}, d_{y}, d_{z}]$ as $\left\{ \overline{p}_{i}=\left( \lfloor \frac{x_{i} - X_{\text{min}}}{d_{x}} \rfloor, \lfloor \frac{y_{i}- Y_{\text{min}}}{d_{y}} \rfloor, \lfloor \frac{z_{i} - Z_{\text{min}} }{d_z} \rfloor \right): i=1, \ldots, N\right\}$, then reorganized into the sparse voxel tensor. The voxel index is the unique integer $\overline{p}_{i}$ and the voxel feature is initialized as the mean of the 3D coordinates and reflection intensities of all the points within the voxel \cite{PartA2_TPAMI,PVRCNN}. The voxel encoder takes the sparse voxel tensor as input for learning the multi-level voxel features $V$ in sparse convolution blocks \cite{submanifold} with downsampling stride $s$. Then the output voxel features are projected to BEV for generating the dense object proposals. We adopt the voxel based networks \cite{VoxelNet,Second} and the regression heads in \cite{Second} as our voxel encoder and RPN.

\subsection{Residual Voxel-to-Point Decoder}\label{sec:res_v2p}
We propose a residual voxel-to-point decoder to provide the full point features with sufficient object positional information. As shown in Figure~\ref{Framework}~(c), we hierarchically aggregate the multi-level voxel features $V$ through the skip connections \cite{Unet,PointNet++} and the staked residual voxel-to-point decoding blocks $\mathcal{D}$. In the block ${\mathcal{D}}^{(l)}$ at level $l$, the current point features $P^{(l)}$ is updated from previous point features $P^{(l+1)}$ and the lateral voxel features $V^{(l)}$ in the residual block \cite{ResNet}. 

\begin{figure}[!tp]%figure* for wider figures.
	\centering
	\includegraphics[height=7.5cm]{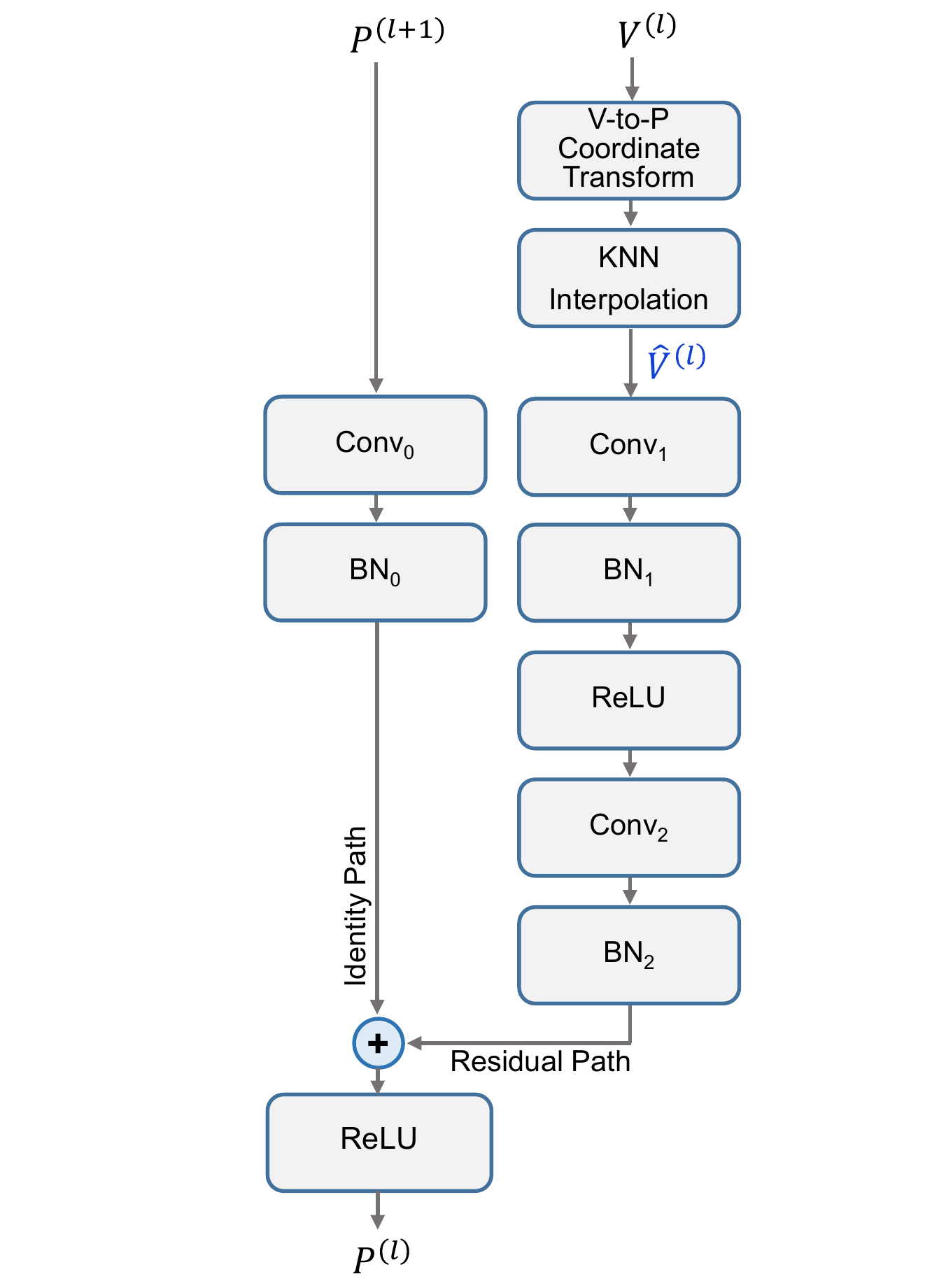}% 
	\caption{Residual voxel-to-point decoding block with Convolution (Conv), Batch Normalization (BN), and ReLU.}
	\vspace{-3mm}
	\label{residual_v2p}
\end{figure} 

Figure~\ref{residual_v2p} presents the decoding block ${\mathcal{D}}^{(l)}$. For each point, considering the neighbors is crucial to the point feature learning \cite{PointNet++} for enlarging the receptive field. The features of the voxels that are in the $K$-Neareast neighbors of the raw point can be aggregated as the voxel-to-point features. The point features $P^{(l+1)}$ on the previous level are used as the identity information while the newly aggregated voxel-to-point features are used as residual information for the enhanced point features $P^{(l)}$. The insight of the design is to conduct the residual learning between the point features from the previous level and the voxel-to-point features for fine-grained feature extraction. We transform the integer voxel index $(v^{(l)}_{x}, v^{(l)}_{y}, {v^{(l)}_{z}})$ back to the raw point coordinate system by representing voxel as point with floating-point ${p'}^{(l)}=({x'}^{(l)}, {y'}^{(l)}, {z'}^{(l)})$ on the voxel center according to the downsampling stride $s^{(l)}$, initial quantization step as $d=[d_{x}, d_{y}, d_{z}]$, and the point cloud boundaries. The voxel-to-point coordinate transformation can be formulated as:
\begin{equation}
	\left\{
		\begin{aligned}
			{{x'}^{(l)}} &= (v^{(l)}_{x} + 0.5) \times d_{x} \times s + X_{\text{min}},\\
			{{y'}^{(l)}} &= (v^{(l)}_{y} + 0.5) \times d_{y} \times s + Y_{\text{min}},\\
			{{z'}^{(l)}} &= (v^{(l)}_{z} + 0.5) \times d_{z} \times s + 
			Z_{\text{min}},
		\end{aligned}
	\right.
\end{equation}
where the offset 0.5 is to centralize the voxel index as the voxel center point. Then we obtain a intermediate point cloud as $\left\{ ({p'}^{(l)}_{i}, V^{(l)}_{i}) : i=1, \ldots, N^{(l)}\right\}$ with points ${p'}^{(l)}$ and the corresponding sparse features $V^{(l)}$ from the voxel encoder. To enhance the previous point features $P^{(l+1)}$ with the intermediate point cloud, we upsample it to be aligned with the resolution of raw points by the interpolation-based feature propagation \cite{PointNet++}. The interpolation is implemented as the inverse distance weighted average $w$ among the $K$-Nearest Neighbors (KNN) in ${p'}^{(l)}$ as:
\begin{align}
% {\hat{V}}^{(l)}_{i} = \sum_{j=1}^{K} \frac{ w_{j}\left(p_{i}\right) V^{(l)}_{j} }{\sum_{j=1}^{K} w_{j}\left(p_{i}\right)}, i=1, \ldots, N,\\
% w_{j}(p_{i}) =\frac{1}{ \left\| {p'}^{(l)}_{j} - p_{i} \right\| }, j=1, \ldots, K,
{\hat{V}}^{(l)}_{i} =& \sum_{k} \frac{ w_{k}\left(p_{i}\right) V^{(l)}_{k} }{\sum_{k} w_{k}\left(p_{i}\right)}, i=1, \ldots, N,\\
& w_{k}(p_{i}) =\frac{1}{ \left\| {p'}^{(l)}_{k} - p_{i} \right\| }.
\end{align}
The decoding block $\mathcal{D}^{(l)}$ finally combines the previous point features $P^{(l+1)}$ and the voxel-to-point feature ${\hat{V}}^{(l)}$ in a manner of residual learning for output $P^{(l)}$. Each voxel has grouped the internal points together in the voxel encoder. From $V^{(4)}$ to $V^{(1)}$, the number of voxels gradually increases. Since the multi-level voxel features are hierarchically aggregated with the residual learning of the point features, the final point features $P^{(1)}$ are gradually enhanced. In the first decoding block $\mathcal{D}^{(4)}$, we can initialize the point feature $P^{(\text{init})}$ to be adapted from the voxel feature $V^{(4)}$ using the interpolation-based feature propagation. The $K$ is set as 3 for computational efficiency.

To explicitly guide the residual voxel-to-point decoder to focus on the 3D structure information of the foreground objects, we use an auxiliary segmentation task for a supervised semantic point feature learning. The last decoding block is followed by a 1-D convolutional block for feature embedding and another one for the auxiliary point-wise semantic segmentation output $S$. The point-wise labels can be generated by determining whether the point is within the annotated box or not. Since the foreground points on the object are less than the background points, especially in the large-scale outdoor scenes, we adopt the focal loss \cite{FocalLoss_tpami2020} with the default settings to deal with such an imbalance.   

\begin{figure}[!tp]%figure* for wider figures.
	\centering
	\includegraphics[width=1.0\linewidth]{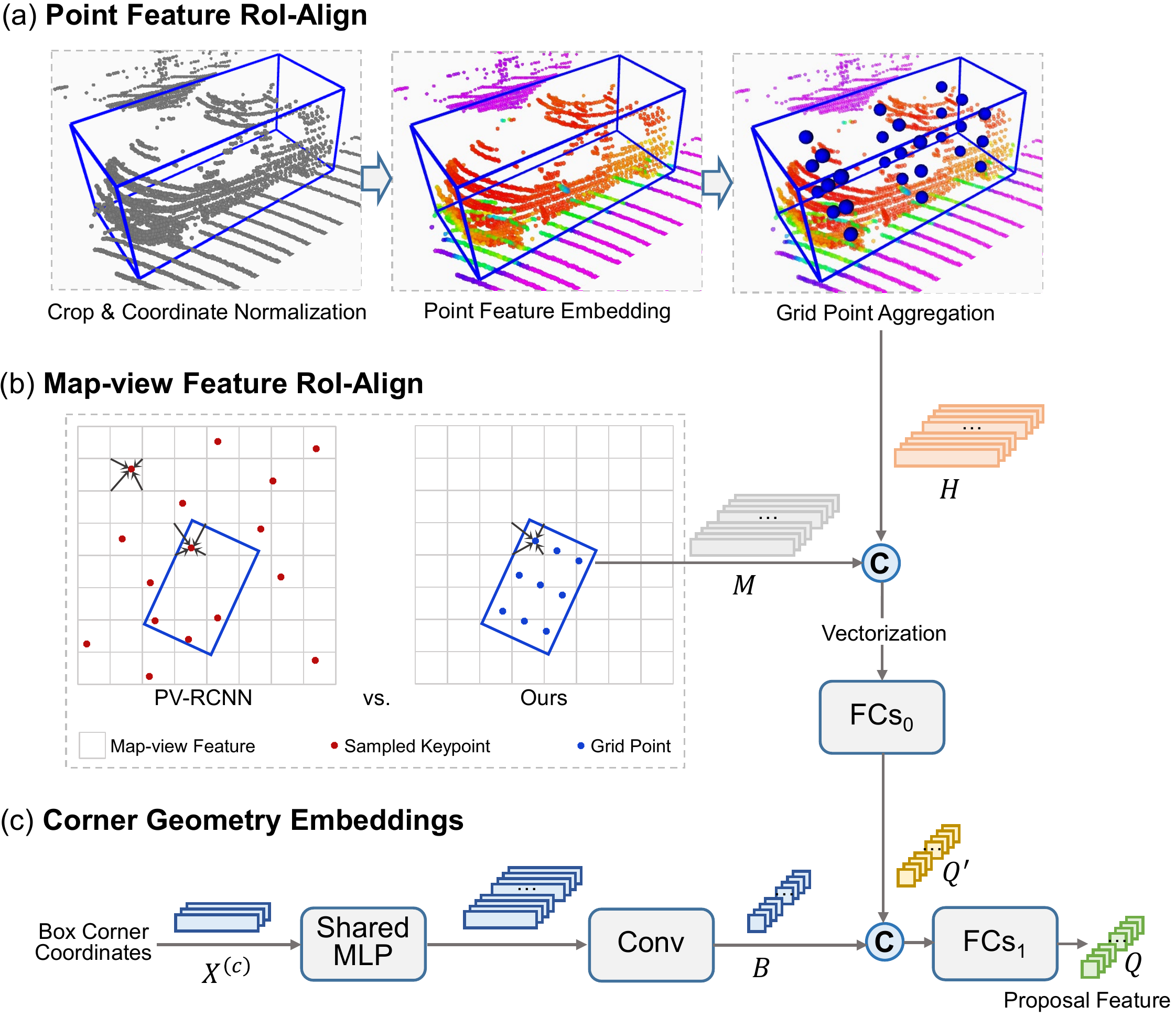}%	
	\vspace{-2mm}
	\caption{Multi-stream RoI pooling. Multi-Layer Perceptron (MLP), Convolution (Conv), Fully Connected layers (FCs).}
	\label{multi_stream_roi_align}
	\vspace{-2mm}
\end{figure}

\subsection{IoU-guided Detection Refinement}\label{sec:iou_guided_refine}

\subsubsection{Multi-stream 3D RoI Pooling}

\textbf{Point Feature RoI-Align.} 
%The point feature RoI-Align operation is performed to accurately perceive the object location with the fused point features $\left\{ (P^{(0)}, S) \right\}$.
The point feature RoI-Align operation is performed to accurately perceive the object location with $P^{(0)}$ and $S$ provided by our voxel-to-point decoder. As shown in Figure \ref{multi_stream_roi_align} (a), the 3D points around a proposal are cropped by the expanded proposal box as the point-stream RoI feature. We add a margin $d_{\text{size}}$ to the size of the box to increase the tolerance of the proposal localization for including more surrounding necessary information of the object. The cropped point cloud is normalized to the relative coordinate system centered on the proposal center and aligned to the proposal rotation angle, which makes it more robust to the rigid transformations \cite{PointRCNN}. Next, we aggregate the point-wise local features with the semantic mask $S$. For each point, the relative coordinates $\tilde{p}_i$ represent the pose information, the depth $\tilde{D}_{i}$ provides the distance information, and the semantic mask $\tilde{S}_i$ makes it sensitive to the foreground and the background information. Note that the symbol ``~$\tilde{~}$~'' indicates the relative coordinate system in the cropped point cloud. The low-dimensional local features $\left\{ ( \tilde{p}_i, \tilde{D}_{i}, \tilde{S}_i  ) \right\}$ are concatenated together and transformed as the expressive high-dimensional feature representation $\tilde{P'}_{i}$ with a Multi-Layer Perceptron (MLP). The point-wise feature is obtained from $\tilde{P'}$ and $\tilde{P^{(0)}}$:
\begin{align}
    \tilde{P'}_{i} = \text{MLP}_{1}([\tilde{p}_{i} \copyright \tilde{D}_{i} \copyright \tilde{S}_{i}]) \label{local_feaure_transform},\\
    \tilde{P}_{i} = \text{MLP}_{2}([ \tilde{P'}_{i} \copyright \tilde{ P^{(0)}}_{i}]),
\end{align}
where ``$\copyright$'' indicates the concatenation. In the third step, we use the grid point aggregation \cite{PVRCNN}. The proposal box is divided into $N_{G} = N_{g} \times N_{g} \times N_{g}$ regularly distributed grid points. Given a grid point $\tilde{g}_{n}$, we group a set of $K_{1}$ neighboring points within radius $r$ for feature aggregation along with set abstraction \cite{PointNet++}:  
\begin{equation}
% {H}_{n}=\max _{k=1,2, \cdots, K_{1}}\left\{\text{MLP}_{3}\left(\left[(\tilde{p}_{k}-\tilde{g}_{n}) \copyright \tilde{P}_{k} \right]\right)\right\}
{H}_{n}=\max _{k} \left\{\text{MLP}_{3}\left[(\tilde{p}_{k}-\tilde{g}_{n}) \copyright \tilde{P}_{k} \right]\right\}
\end{equation}
where ${\left\| {\tilde{p}_{k} - \tilde{g}_{n}} \right\|} < {r}$. The $C_{h}$-dimensional RoI-Aligned point feature can be formulated with all the grid points:
\begin{align}
    {H} = [{H}_{1}, \cdots, {H}_{n}, \cdots, {H}_{N_{G}}] \in \mathbb{R}^{C_{h} \times N_{G}}.
\end{align}
We set the multiple group radii in the set abstraction for multi-scale information aggregation \cite{PointNet++}, and determine the box size margin $d_\text{size}$ as the group radius.

\textbf{Map-view Feature RoI-Align.}
% B 代表 BEV Map-view Feature
Since the receptive field of the map-view feature in RPN increases in the hierarchical voxel encoder blocks, the map-view feature discards the details of objects but retains the structural information of objects. Thus, we use the grid points instead of the sampled keypoints in \cite{PVRCNN} to extract the compact map-view feature in the regions of objects. The local grid points $\tilde{g}$ is first transformed back to the global coordinate system in the raw point cloud. Then we project them onto BEV. The coordinates $(g^{(bev)}_{x},g^{(bev)}_{y})$ of the projected grid points on the map-view feature can be calculated by grid point global coordinates $g$, down-sample stride $s^{(bev)}$ of map-view feature, quantization $d$, and point cloud boundary as: 
\begin{equation}
	\left\{
		\begin{aligned}
			{{g}^{(bev)}_{x}} &= (g_{x} - X_{\text{min}}) / (d_{x} \times s^{(bev)}),\\
			{{g}^{(bev)}_{y}} &= (g_{y} - Y_{\text{min}}) / (d_{y} \times s^{(bev)}).\\
		\end{aligned}
	\right.
\end{equation}
To avoid the quantization errors as introduced in \cite{fasterrcnn,Maskrcnn}, we directly keep the floating-point coordinates $g^{(bev)}$ instead of rounding them. Given a projected grid point $g^{(bev)}_{n}$, we use the bilinear interpolation to compute its map-view feature $M_{n}$ from its four nearest integer neighbors, which is clearly illustrated in Figure \ref{multi_stream_roi_align} (b). The $C_{m}$-dimensional map-view RoI feature $M$ aligned by $N_{G}$ grid points can be formulated as:
\begin{align}
    M = [M_{1}, \cdots, M_{n}, \cdots, M_{N_{G}}] \in \mathbb{R}^{C_{m} \times N_{G}}.
\end{align}

\textbf{Corner Geometry Embeddings.}
The eight corners of the box are not usually included in the raw points, but they are closely related to the localization accuracy. Most methods only focus on the RoI features and ignore the geometry information of the boxes. The Corner Geometry Embeddings (CGEs) include the box geometry information in addition to the RoI features. As illustrated in Figure~\ref{multi_stream_roi_align} (c), the eight corners' coordinates $X^{(c)} \in \mathbb{R}^{3 \times 8}$ are projected into a $C_{b}'$-dimensional space, ensuring that the low-dimensional coordinates of corners are not overwhelmed by the high-dimensional RoI features. Secondly, a 1-D convolution with kernel $\theta \in \mathbb{R}^{C_{b} \times {C_{b}'} \times 8}$ is applied to the eight corners for the final CGEs $B \in \mathbb{R}^{C_{b} \times 1}$.

\textbf{Multi-stream Feature Aggregation.}
As shown in Figure \ref{multi_stream_roi_align}, the RoI-Aligned point feature ${H}$ and map-view feature $M$ are first concatenated together, then vectorized to be fused as $Q'$ in a set of Fully Connected layers (FCs). Another set of FCs is used to transform the concatenation of the CGEs $B$ and $Q'$ as the final proposal feature $Q$ for the detection branches, including a classification, a box regression, and an IoU estimation branch. All the three parallel branches are implemented in the same structure, consisting of a set of FCs for feature embedding and another fully connected layer with a different number of neurons adapting to the dimension of the output.

\subsubsection{3D IoU Alignment}
The IoU misalignment is caused by the proposal box refinement. Since the estimated IoUs are regressed from RoI-Aligned features using the proposal boxes, they are aligned to the proposal boxes, not the refined proposal boxes. The estimated IoUs from the proposal boxes cannot be used as the localization confidence for the refined proposal boxes in NMS. To align the estimated IoUs to the refined boxes, we replace the proposal boxes in the 3D RoI pooling with the refined proposal boxes in one more inference stage without training. Then the computed IoUs are aligned to the refined proposal boxes and used as the localization confidence for the refined proposal boxes de-redundancy in NMS.

%We can train one more 3D IoU alignment stage by copying the refinement stage, then updating the proposal boxes with the refinement proposal boxes to train another IoU alignment stage, and aligning the IoU in the inference stage. We discuss the outcome of one more 3D IoU alignment stage in the experiment.  

\subsection{Loss Function}\label{sec:loss_func}
Our network is optimized by a multi-task loss $\mathcal{L}_{\text{total}}$ as:
\begin{align}
    \mathcal{L}_{\text{total}} = \alpha_{1}\mathcal{L}_{\text{rpn}} + \alpha_{2}\mathcal{L}_{\text{seg}}  + \alpha_{3}\mathcal{L}_{\text{refine}}, \label{loss_total} 
\end{align}
where the coefficients $\alpha_{1}$, $\alpha_{2}$, and $\alpha_{3}$ are set to 1.0, 4.0, and 1.0 to balance the RPN loss $\mathcal{L}_{\text{rpn}}$ in \cite{Second}, the point segmentation loss $\mathcal{L}_{\text{seg}}$ to supervise the semantic point feature learning, and the detection refinement loss $\mathcal{L}_{\text{refine}}$, respectively. The $\mathcal{L}_{\text{refine}}$ can be further formulated as :
\begin{align}
    \mathcal{L}_{\text{refine}} = \mathcal{L}_{\text{cls}} + \mathcal{L}_{\text{reg}} \label{loss_total} + \mathcal{L}_{\text{iou}}.
\end{align}
The classification loss $\mathcal{L}_{\text{cls}}$ is computed using the binary cross entropy loss. %between the foreground object classification labels $c$ and the predicted classification score $\hat{c}$ on the $N$ proposal samples sampled for the refinement stage as:
% \begin{align}
%     \mathcal{L}_{\text {cls}}= \frac{1}{N} \sum_{i} \mathcal{L}_{\text {bce}}\left(c_{i}, \hat{c_{i}} \right).
% \end{align}
The box regression loss $\mathcal{L}_{\text{reg}}$ and the IoU estimation loss $\mathcal{L}_{\text{iou}}$ are both computed using the smooth-L1 loss on the $N_{\text{reg}}$ proposals with $\text{IoU} \geqslant \theta_{\text{reg}}$ in a manner as: 
%The the box regression loss $\mathcal{L}_{\text{reg}}$ and IoU estimation loss $\mathcal{L}_{\text{iou}}$ are both computed by the smooth-L1 loss in a manner as: 
\begin{align}
    \mathcal{L}= \frac{1}{N_{\text{reg}}} \sum_{i}  \left[ \text{IoU}_{i} \geqslant \theta_{\text{reg}}\right] \mathcal{L}_{\text{smooth-L1}} \left( a_{i}, \hat{a_{i}} \right), \label{eq_reg_loss}
\end{align}
where $a_{i}$ and $\hat{ a_{i} }$ denote the target and prediction of specific item (\textit{i.e.}, IoU and the residuals of center coordinates, size, orientation) for the $i$-th proposal. The Iverson bracket indicator function $[\text{IoU}_{i} \geqslant \theta_\text{reg}]$ sets as 1 when $\text{IoU}_i \geqslant \theta_{\text{reg}}$, otherwise, it sets as 0. %$N_{\text{reg}}$ is the number of proposals that satisfy the condition $\text{IoU}_i \geqslant \theta_{\text{reg}} $.

\section{Experiments}
\subsection{Datasets}
\textbf{KITTI Dataset.}
The KITTI \cite{KITTIDataset} is one of the most widely used 3D object detection datasets for autonomous driving. It contains 7481 training samples and 7518 testing samples, and annotates objects in the camera Field of Vision (FOV). We follow the common practice to divide the training samples as \textit{train} split set (3712 samples) and the \textit{val} split set (3769 samples) \cite{MV3D,Second,PVRCNN}. For submitting the results on the \textit{test} set to the online benchmark, the training samples are randomly divided into two sets at a ratio of $4:1$ for training and validation following \cite{PVRCNN,STD}.

\textbf{Waymo Open Dataset.}
The newly released Waymo Open Dataset (WOD) \cite{waymo_dataset} is currently the largest public dataset for autonomous driving, including $\sim$158K point cloud training samples and $\sim$40K point cloud validation samples. Different from KITTI, the WOD provides object annotations in the full \ang{360} fields. To further verify the effectiveness of our method, we also evaluate the performance of our method on the more challenging dataset WOD. 

\vspace{-1mm}
\subsection{Implementation Details}
\vspace{-1mm}
\textbf{Voxelization.}
Since the KITTI only provides the annotations in FOV, we set the quantization step $d$ as $(0.05, 0.05, 0.1)$ meters to voxelize point cloud within the range of $\left[ 0, 70.4 \right], \left[ -40, 40 \right], \left[ -3, 1 \right]$ meters in $X$, $Y$, $Z$-axis. For WOD, the range of point cloud is $\left[ -75.2, 75.2 \right], \left[ -75.2, 75.2 \right], \left[ -2, 4 \right]$ meters for $X$, $Y$, $Z$-axis, and the quantization step $d$ is $(0.1, 0.1, 0.15)$ meters. 

\textbf{Network Architectures.}
The architecture of our voxel encoder follows the design in \cite{Second,PVRCNN}. It downsamples the voxel volumes with $1 \times$, $2 \times $ $4 \times$, $8 \times$ while it increases the feature dimension as 16, 32, 64, 128. The RPN head is adopted from the \cite{Second} for region proposal. The residual voxel-to-point decoder gradually aggregates the different number of voxels from the different voxel feature levels to the raw point clouds with the output feature dimension as 256, 192, 160, 128 and 128 for $P^{(4)}$, $P^{(3)}$, $P^{(2)}$, $P^{(1)}$ and $P^{(0)}$, respectively. For each proposal in the detection refinement stage, $N_{G} = 6 \times 6 \times 6$ grid points are generated to aggregate the cropped point clouds with multiple group radii $(0.8, 1.6)$ meters. The feature dimension $C_{h}$, $C_{m}$ and $C_{b}$ are set as 128 for RoI-Aligned point features, RoI-Aligned map-view features and corner geometry embeddings.

\begin{figure}[!t]%figure* for wider figures.
	\centering
	\includegraphics[width=0.5\linewidth]{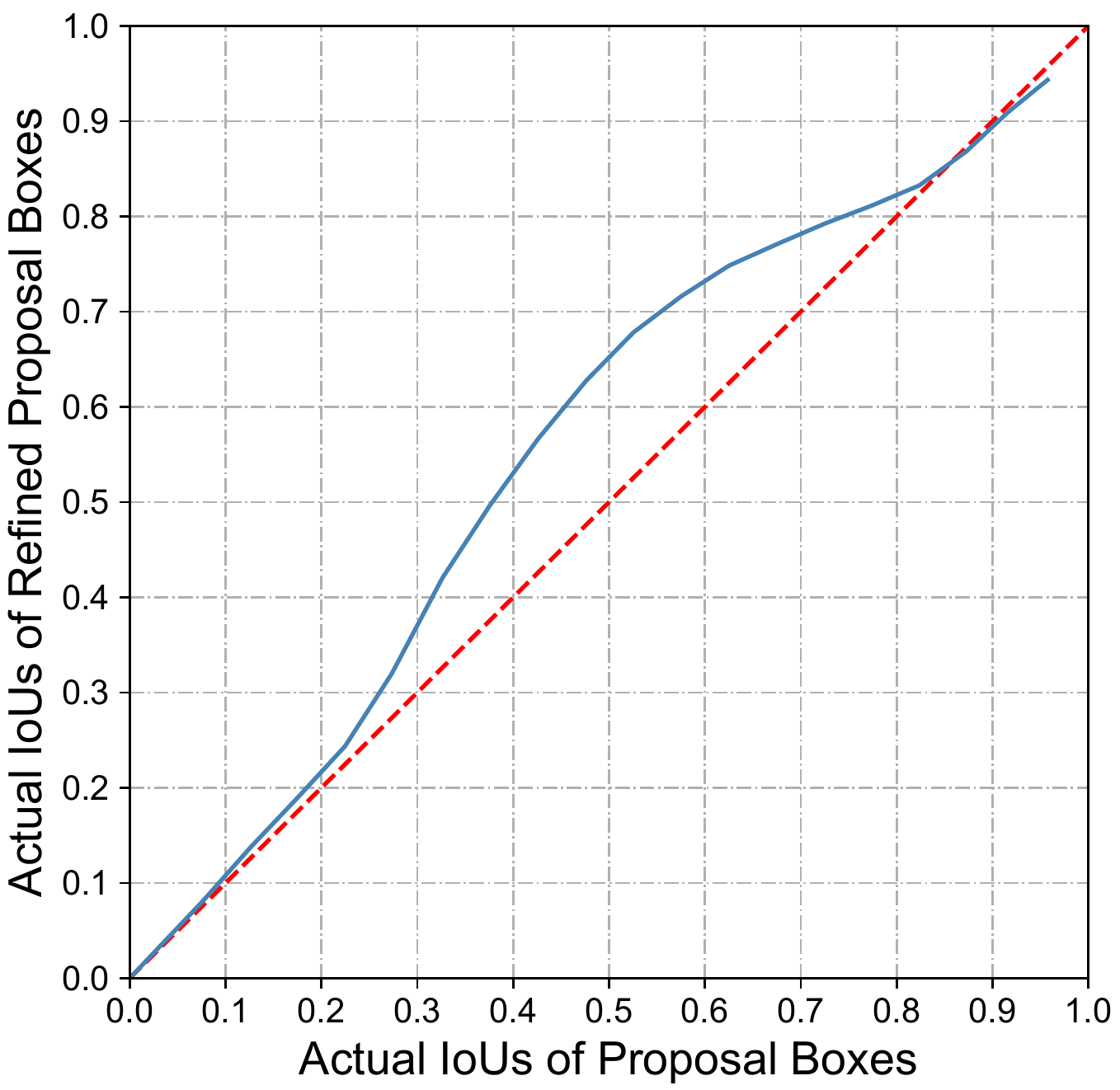}%
	\vspace{-4mm}
	\caption{IoU curve of proposal boxes and refined boxes.}
	\vspace{-2mm}
	\label{inputiou_vs_outputiou}
\end{figure}
%The top 100 proposal boxes and the corresponding refined boxes of each KITTI \textit{val} point cloud sample are counted.

\begin{table}[!t]
\caption{Detection performance comparison of the architectures without and with detection refinement. The classification score is used as the detection confidence following PV-RCNN\cite{PVRCNN} to exclude the effects of IoU alignment.}
\vspace{-3mm}
\begin{center}
\setlength{\tabcolsep}{2.0mm}
\resizebox{0.35\textwidth}{!}{
\begin{tabular}{c|ccc}
\toprule
\multirow{2}{*}{Detection Refinement} & \multicolumn{3}{c}{3D AP} \\
                              & Easy   & Mod.   & Hard   \\
\midrule
$\times$                  & 88.98  & 80.85  & 78.79  \\
$\surd$                    & \textbf{92.69}  & \textbf{85.38}  & \textbf{83.23}  \\
\bottomrule
\end{tabular}}
\end{center}
\label{AP_1satge_vs_2stages}
% \vspace{-3mm}
\end{table}

\textbf{Training.}
Our model is trained from scratch in an end-to-end manner with the AdamW optimizer \cite{AdamW} and one-cycle policy \cite{one_cycle_lr} with LR 0.01, division factor 10, momentum ranges from 0.95 to 0.85, weight decay 0.01. A batch of 8 or 48 random point cloud samples is trained on 4 or 16 Tesla V100 GPUs with 80 or 30 epochs for KITTI and WOD, respectively. For the detection refinement stage, we sample 128 proposals from the RPN as the training samples. The foreground IoU threshold $\theta_{\text{H}}$ is set as 0.75, the background IoU threshold $\theta_{\text{L}}$ is set as 0.25 for classification branch. The threshold $\theta_{\text{reg}}$ mentioned in Equation~\ref{eq_reg_loss} is empirically set as 0.55 to select approximately half of the sampled proposals for box regression and IoU estimation training, which follows \cite{PVRCNN,PointRCNN,STD}. To avoid overfitting, we employ four commonly used data augmentation strategies: ground truth sampling \cite{Second}, random flipping along the $X$-axis, global scaling with a random scaling factor in $[0.95, 1.05]$, global rotation around the $Z$-axis with a random angle in $[- \frac{\pi}{4}, \frac{\pi}{4} ]$. 

\textbf{Inference.} 
We perform the NMS on the RPN proposals with IoU threshold 0.85 to take the top-100 proposals as the input of the detection refinement stage. After refining the top-100 proposals, we align the estimated IoUs to the refined boxes. %with the proposed 3D IoU alignment method. %Finally, we select the positive detection results with their classification score~\textgreater~0.3, and remove the redundant boxes by the aligned-IoUs based NMS with IoU threshold 0.1.

\subsection{Ablation Study}
We conduct an extensive ablation study to analyze each component in our method. All models are trained with the same settings on the KITTI \textit{train} split set and evaluated with 3D average precision (AP) from 40 recall positions for the car class on the KITTI \textit{val} split set \cite{SASSD, PVRCNN, 3DSSD}. The best results are in bold. Objects are marked as three detection difficulties (easy, moderate, and hard) by KITTI, depending on their size, occlusion level, and truncation of 3D boxes.

\begin{figure}[!t]%figure* for wider figures.
	\centering
	\includegraphics[width=0.8\linewidth]{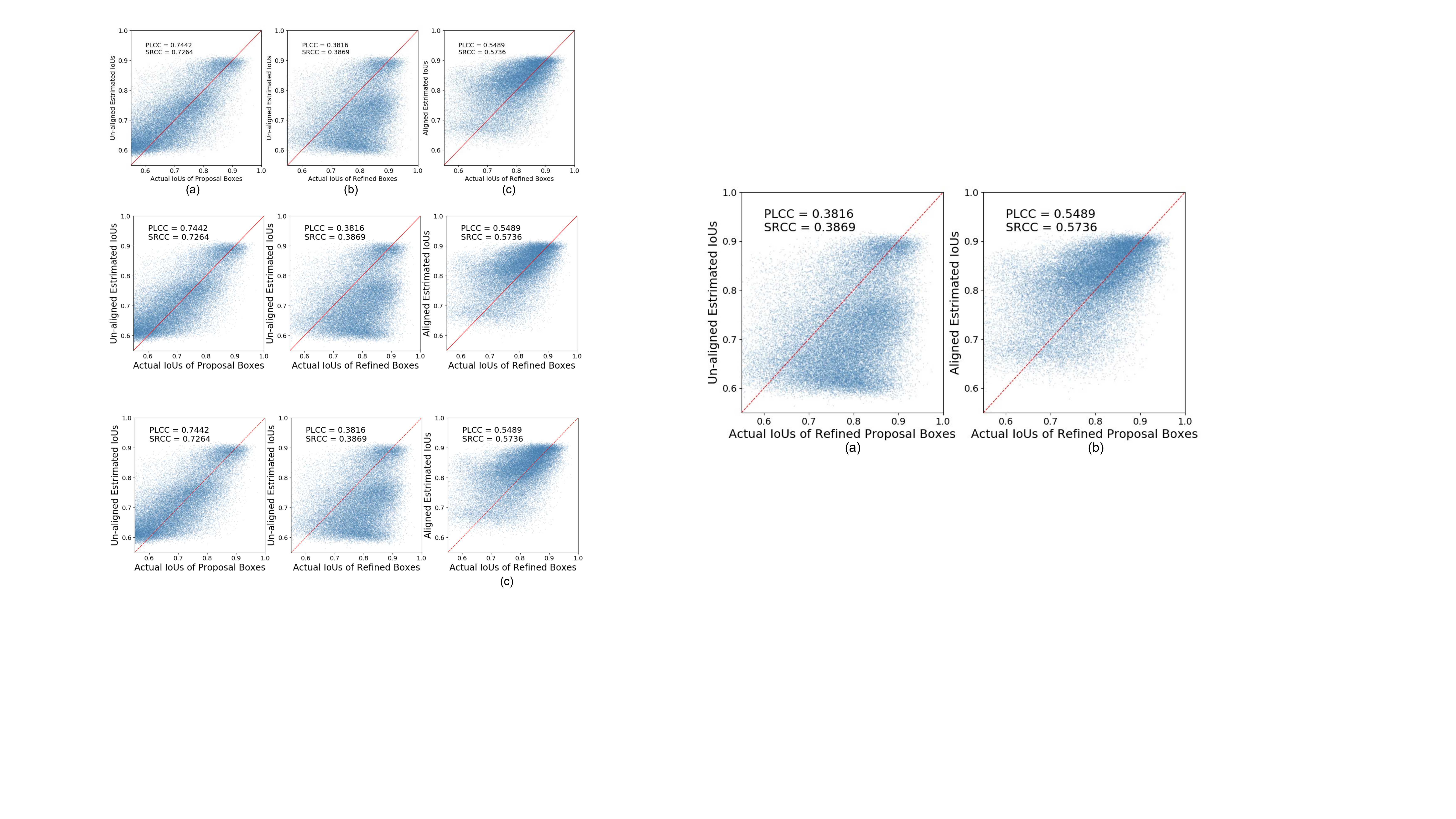}%
	\vspace{-4.5mm}
	\caption{Correlation between the estimated IoUs and the actual IoUs. Pearson’s Linear Correlation Coefficient (PLCC) and Spearman’s Rank order Correlation Coefficient (SRCC) describe the linear and the order correlation. Higher, better.}
	\label{correlation_scatter}
% 	\vspace{-5mm}
\end{figure}

\begin{table}[!t]
\caption{Comparison of different localization confidences.}
\vspace{-3mm}
\label{AP_different_detection_confidence}
\small
\setlength{\tabcolsep}{0.6mm}
\resizebox{0.47\textwidth}{!}{
\begin{tabular}{c|c|ccc}
\toprule
\multirow{2}{*}{Tag} & \multirow{2}{*}{Localization Confidence} & \multicolumn{3}{c}{3D AP} \\
                     &                                       & Easy    & Mod.   & Hard   \\
\midrule
A                    & Classification score                 & 92.69   & 85.38  & 83.23  \\
B                    & Un-aligned IoUs                      & 92.51   & 84.95  & 82.89  \\
C                    & Aligned IoUs                         & \textbf{93.09}   & 85.54  & 83.38  \\
D                    & Aligned IoUs$\times$Classification score    & 93.00      & \textbf{85.61}  & \textbf{83.43}\\ 
\bottomrule
\end{tabular}
}
% \vspace{-3mm}
\end{table}

\textbf{Detection Refinement Stage.} % or somewhat like the existence of the IoU mismatch
% AP 对比表 + inputiou vs outputiou
Table~\ref{AP_1satge_vs_2stages} shows that the detection refinement stage can significantly improve the performance. We plot the actual IoU changes of the proposal boxes and the refined proposal boxes in Figure~\ref{inputiou_vs_outputiou}. It shows that the actual IoUs of the refined proposal boxes become higher than those of the proposal boxes, indicating the improvements in localization accuracy. %This implies the effectiveness of the detection refinement.

\textbf{3D IoU Alignment.}
We use four sets of localization confidence as shown in Table \ref{AP_different_detection_confidence}. When we directly use the un-aligned IoUs estimated from the proposal boxes as the localization confidence, the localization accuracy is worse than that based on the classification score. To address the IoU misalignment, we do one more inference to estimate the IoUs from the RoI features aligned with the refined proposal boxes. The aligned IoUs guided NMS achieves higher 3D AP than the first two in Table~\ref{AP_different_detection_confidence}. As shown in Figure~\ref{correlation_scatter}, compared with the un-aligned IoUs, the aligned IoUs become more correlated to the actual IoUs of the refined proposal boxes. Besides, when we multiply the aligned IoUs and the classification score following STD \cite{STD}, the detection performance can be further improved.

\textbf{3D RoI Pooling.}
Table \ref{multistream_RoI_Pooling_AP} shows the effectiveness of each individual feature stream. When the point features are removed, the detection performance drops significantly, which implies that the fine-grained point features provide additional information to improve the localization accuracy. For the moderate and hard objects with much more sparse points, the 3D AP drops by 2.73\% and 3.19\%. As shown in the last two rows of Table \ref{multistream_RoI_Pooling_AP}, the map-view features and the corner geometry embeddings also contribute to the performance gains by providing the object structure information and the box corner information. As we mentioned, the map-view feature loses the detailed pose information that is preserved in the points. Thus, the point features contribute more than the map-view features to the performance improvement.

\begin{table}[t!]
\caption{Effects of each feature stream in 3D RoI-Pooling. Point, Map, and CGEs denote the RoI-Aligned point features, the RoI-Aligned map-view features, and the corner geometry embeddings.}
\vspace{-3mm}
\label{multistream_RoI_Pooling_AP}
\setlength{\tabcolsep}{1.6mm}
\begin{center}
\resizebox{0.35\textwidth}{!}{
\begin{tabular}{ccc|ccc}
\toprule
\multirow{2}{*}{Point} & \multirow{2}{*}{Map} & \multirow{2}{*}{CGEs} &       & 3D AP &       \\
          &          &          & Easy  & Mod.  & Hard  \\
\midrule
$\surd$         & $\surd$   & $\surd$       & \textbf{93.00} & \textbf{85.61} & \textbf{83.43} \\
$\times$        & $\surd$   & $\surd$       & 92.38 & 82.88 & 80.24 \\
$\surd$         & $\times$  & $\surd$       & 92.78 & 85.32 & 83.15 \\
$\surd$         & $\surd$   & $\times$      & 92.91 & 85.41 & 83.21 \\
\bottomrule
\end{tabular}}
\end{center}
% \vspace{-2mm}
\end{table}

% \noindent
\textbf{Supervised Point Feature Learning.}
We train a model with our best model structure but learning point features in an unsupervised manner. Table \ref{AP_sup_vs_unsup} shows that the point segmentation supervision brings performance improvements of 0.21\%, 0.27\%, and 0.33\% on 3D AP, indicating that the semantic information is useful in the point feature learning. %Besides, the quantitative results in Figure x show that the point cloud segmentation task can also be learned in our 3D object detection network even without segmentation supervision. The interesting results indicate that the point cloud segmentation and 3D object detection can be jointly promoted.

\begin{table}[!t]
\caption{Performance comparison of the point feature learning with and without segmentation guidance.}
\vspace{-4mm}
\begin{center}
\setlength{\tabcolsep}{2.0mm}
\resizebox{0.4\textwidth}{!}{
\begin{tabular}{c|ccc}
\toprule
\multirow{2}{*}{Segmentation Supervision} & \multicolumn{3}{c}{3D AP} \\
                              & Easy   & Mod.   & Hard   \\
\midrule
$\times$                  & 92.79  & 85.34  & 83.10  \\
$\surd$                   & \textbf{93.00} & \textbf{85.61} & \textbf{83.43}  \\
\bottomrule
\end{tabular}}
\end{center}
\label{AP_sup_vs_unsup}
% \vspace{-4mm}
\end{table}

\begin{figure*}[t!]%figure* for wider figures.
	\centering
	\includegraphics[width=0.8\linewidth]{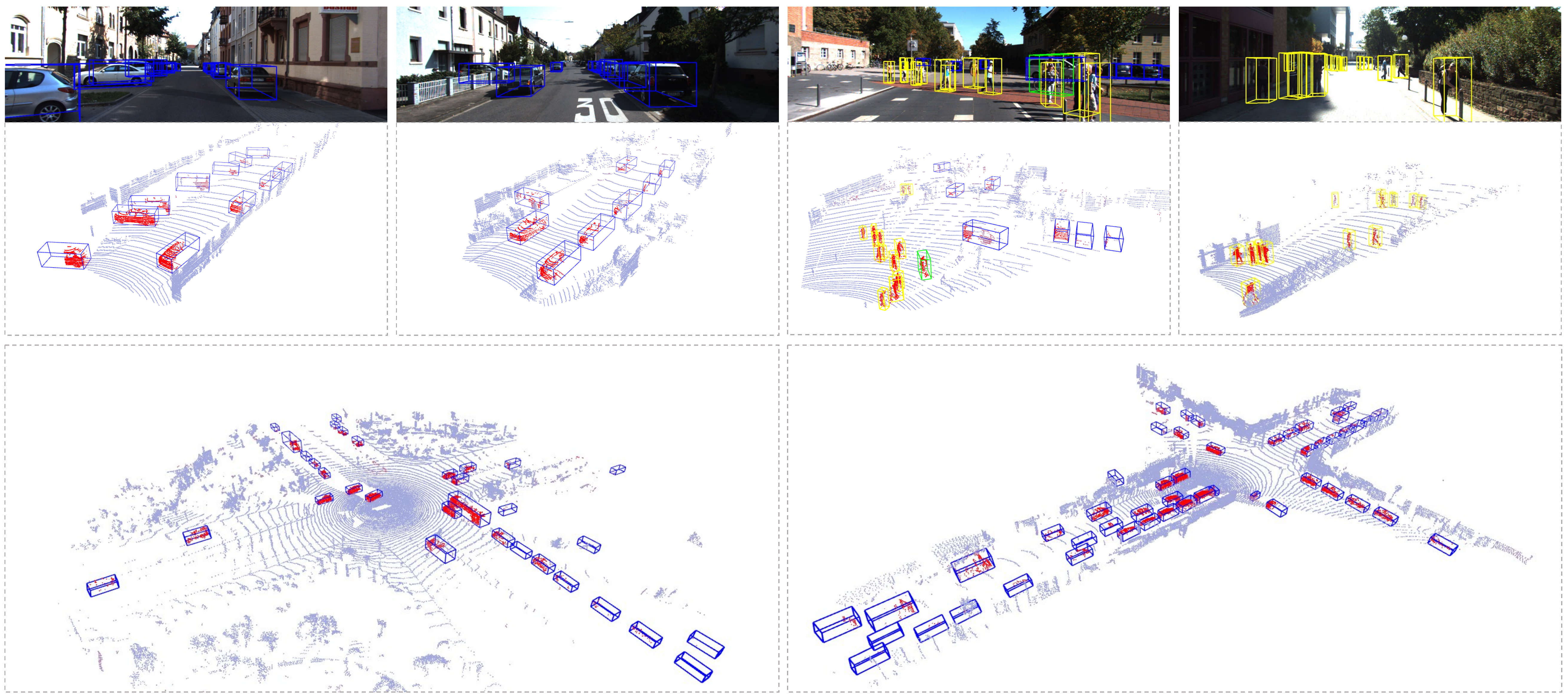}
	\vspace{-3mm}
	\caption{Qualitative results on KITTI (FOV) and Waymo (\ang{360}). Detected cars, cyclists, and pedestrians with blue, green, and yellow boxes, respectively. The color in point cloud shows segmentation results. Foreground in red. Background in gray.}
	\label{qualitative_results}
\end{figure*}

\begin{table}[!t]
% \caption{Performance comparison on car class of the KITTI \textit{test} set by submitting to the official test server. The top-2 results are in bold. ``L'' and ``C'' indicate the LiDAR and camera, respectively.}
\caption{Performance comparison on car class of the KITTI \textit{test} set. The top-2 results are in bold. ``L'' and ``C'' indicate the LiDAR and camera, respectively.}
\vspace{-3mm}
\label{ap_kitti_test}
\small
\begin{center}
\resizebox{0.47\textwidth}{!}{
\begin{tabular}{c|c|ccc}
\toprule
\multirow{2}{*}{Method} & \multirow{2}{*}{Modality} & \multicolumn{3}{c}{3D AP} \\
                        &                           & Easy    & Mod.   & Hard   \\
\midrule
MV3D\cite{MV3D}         & L+C                       & 74.97   & 63.63  & 54.00  \\
AVOD\cite{AVOD}         & L+C                       & 83.07   & 71.76  & 65.73  \\
Conti-Fuse\cite{Contifuse}  & L+C                       & 83.68   & 68.78  & 61.67  \\
F-PointNet\cite{F-PointNets}    & L+C                       & 82.19   & 69.79  & 60.59  \\
UberATG-MMF\cite{MMF}       & L+C                       & 88.40   & 77.43  & 70.22  \\
PointPainting\cite{pointpainting}  & L+C                    & 82.11 	& 71.70  & 67.08  \\
PI-RCNN\cite{PIRCNN}        & L+C                       & 84.37   & 74.82  & 70.03  \\
%3D-CVF\cite{3DCVF}                  & L+C                       & 89.20   & 80.05  & 73.11  \\
\midrule
VoxelNet\cite{VoxelNet}     & L                         & 77.47   & 65.11  & 57.73  \\
SECOND\cite{Second}         & L                         & 83.34   & 72.55  & 65.82  \\
PointPillars\cite{PointPillars} & L                         & 82.58   & 74.31  & 68.99  \\
PointRCNN\cite{PointRCNN}   & L                         & 86.96   & 75.64  & 70.70  \\
Fast PointRCNN\cite{Fast-PointRCNN}  & L                         & 85.29   & 77.40  & 70.24  \\
STD\cite{STD}       & L                         & 87.95   & 79.71  & 75.09  \\
TANet\cite{tanet_AAAI}       & L                         & 84.39   & 75.94  & 68.82  \\
HotSpotNet\cite{Hotspots(ECCV2020)}   & L                         & 87.60   & 78.31  & 73.34  \\
Part-A2\cite{PartA2_TPAMI}      & L                         & 87.81   & 78.49  & 73.51  \\
Associate-3Ddet\cite{Associate-3Ddet}  & L                         & 85.99   & 77.40  & 70.53  \\
SERCNN\cite{sercnn_cvpr2020} & L                        & 87.74   & 78.96  & 74.30  \\
Point-GNN\cite{PointGNN}    & L                         & 88.33   & 79.47  & 72.29  \\
3DSSD\cite{3DSSD}           & L                         & 88.36   & 79.57  & 74.55  \\
% SA-SSD\cite{SASSD}        & L                         & 88.75   & 79.79  & 74.16  \\
PV-RCNN\cite{PVRCNN}        & L                         & \textbf{90.25}   & \textbf{81.43}  & \textbf{76.82}  \\
%CIA-SSD                 & L                         & 89.59   & 80.28  & 72.87  \\
\textbf{Ours}            & L                         & \textbf{88.53}   & \textbf{81.58}  & \textbf{77.37}  \\
% Ours                    & L                         & \textbf{88.96}   & \textbf{81.51}  & \textbf{77.27}  \\
\bottomrule
\end{tabular}}
\end{center}
\vspace{-5mm}
\end{table}

\subsection{Results on KITTI Dataset}
% 在前面补充mod、hard的特点。
% 分析，尤其是在mod和hard上的提升。
\textbf{Evaluation Metric.} We evaluate our method on the KITTI \textit{test} set following the common practice to report the 3D AP calculated from 40 recall positions. The official benchmark sets the IoU thresholds for cars, cyclists, and pedestrians as 0.7, 0.5, and 0.5, respectively. 

We report the 3D AP on the KITTI \textit{test} set in Table~\ref{ap_kitti_test} and Table~\ref{ap_kitti_test_cyc_ped}. Table \ref{ap_kitti_test} shows the performance comparison on the commonly used car class. Our method that only uses the point cloud from LiDAR outperforms both the LiDAR + Camera based and the LiDAR only based existing methods, especially on the challenging moderate and hard objects with much more sparse points. Compared with PV-RCNN \cite{PVRCNN}, our method increases by 0.15\% and 0.55\% on the moderate and hard objects. Note that the performance on the moderate difficulty is used as the ranking criterion in the official KITTI leaderboard. Table \ref{ap_kitti_test_cyc_ped} shows the performance comparison between our method and other methods on the KITTI $\textit{test}$ set for cyclist and pedestrian  classes, respectively. Our method also achieves state-of-the-art performance against the existing methods. The qualitative results are shown at the top of Figure~\ref{qualitative_results}. 

\begin{table}[!t]
\caption{Performance comparison for the classes of cyclist and pedestrian on the KITTI \textit{test} set by submitting to the official test server. The top-2 results are in bold.}
\vspace{-3mm}
\label{ap_kitti_test_cyc_ped}
\small
\setlength{\tabcolsep}{0.6mm}
\resizebox{0.47\textwidth}{!}{
\begin{tabular}{c|ccc|ccc}
\toprule
\multirow{2}{*}{Method} & \multicolumn{3}{c|}{Cyclist} & \multicolumn{3}{c}{Pedestrian}  \\
                        & Easy    & Mod.    & Hard  & Easy    & Mod.    & Hard      \\
\midrule
AVOD-FPN\cite{AVOD}                & 63.76   & 50.55   & 44.93 & 50.46   & 42.27   & 39.04 \\
BirdNet+\cite{birdnetplus}   	& 67.38    &47.72 	&42.89   &37.99 	&31.46  &29.46\\
SCNet\cite{SCNet}                       & 67.98 & 50.79  & 45.15  & 47.83 	& 38.66	& 35.70  \\ 
F-PointNet\cite{F-PointNets}              & 72.27   & 56.12   & 49.01  & 50.53   & 42.15   & 38.08    \\
PointRCNN\cite{PointRCNN}               & 74.96   & 58.82   & 52.53 & 47.98   & 39.37   & 36.01     \\
PointPillars\cite{PointPillars}            & 77.10    & 58.65   & 51.92   & 51.45   & 41.92   & 38.89     \\
STD\cite{STD}                     & \textbf{78.69}   & 61.59   & 55.30     & \textbf{53.29} & \textbf{42.47} & 38.35 \\
PointPainting\cite{pointpainting}           & 77.63   & 63.18   & 55.89  & 50.32   & 40.97   & 37.87    \\
SemanticVoxels\cite{semanticvoxels}           & N/A     & N/A   & N/A   &50.90  &42.19  &\textbf{39.52}\\
% TANet\cite{tanet_AAAI}                   & 75.70    & 59.44   & 52.53     \\
PV-RCNN\cite{PVRCNN}                  & 78.60    & \textbf{63.71} & \textbf{57.65} & N/A     & N/A   & N/A \\  
\textbf{Ours}                    & \textbf{81.49}   & \textbf{63.41}   & \textbf{56.40}   & \textbf{51.80} & \textbf{43.28} & \textbf{40.71}\\
\midrule
\end{tabular}}
\end{table}

\begin{table}[!t]
\caption{Performance on the Waymo Open Dataset validation for car detection. The top-1 results are in bold.}
\vspace{-2mm}
\label{ap_waymo_val}
\small
\setlength{\tabcolsep}{0.6mm}
\resizebox{0.48\textwidth}{!}{

\begin{tabular}{c|c|cccc}
\toprule
Difficulty                 & Method       & overall & {[}0, 30)m & {[}30, 50)m & {[}50, Inf)m \\
\midrule
\multirow{14}{*}{LEVEL\_1} &              & \multicolumn{4}{c}{3D AP}                          \\
\cline{2-6}
                           & PointPillars\cite{PointPillars} & 56.62   & 81.01      & 51.75       & 27.94        \\
                           & MVF\cite{MVF}          & 62.93   & 86.30       & 60.02       & 36.02        \\
                           & Pillar-OD\cite{pillar-od}    & 69.80    & 88.53      & 66.50        & 42.93        \\
                           & PV-RCNN\cite{PVRCNN}      & 70.30    & 91.92      & 69.21       & 42.17        \\
                           & \textbf{Ours}         & \textbf{77.24}   & \textbf{93.23}      & \textbf{76.21}       & \textbf{55.79}        \\
\cline{2-6}
                           & \textit{Improvement}  & +6.94   & +1.31     & +7.00     & +12.86\\
\cline{2-6}
                           &              & \multicolumn{4}{c}{BEV AP}                        \\
\cline{2-6}
                           & PointPillars\cite{PointPillars} & 75.57   & 92.10       & 74.06       & 55.47        \\
                           & MVF\cite{MVF}          & 80.40    & 93.59      & 79.21       & 63.09        \\
                           & Pillar-OD\cite{pillar-od}    & 87.11   & 95.78      & 84.74       & 72.12        \\
                           & PV-RCNN\cite{PVRCNN}      & 82.96   & 97.35      & 82.99       & 64.97        \\
                           & \textbf{Ours}         &\textbf{88.93}       & \textbf{98.05}          & \textbf{88.25}          & \textbf{79.19}            \\
\cline{2-6}
                           & \textit{Improvement}  & +1.82   & +0.70     & +3.51     & +7.07\\
\midrule
\multirow{6}{*}{LEVEL\_2}  &              & \multicolumn{4}{c}{3D AP}                          \\
\cline{2-6}
                           & PV-RCNN\cite{PVRCNN}      & 65.36   & 91.58      & 65.13       & 36.46        \\
                           & \textbf{Ours}         & \textbf{69.77}   & \textbf{92.53}      & \textbf{70.09}       & \textbf{43.96}        \\
\cline{2-6}
                           & \textit{Improvement}  & +4.41  & +0.95 & +4.96   & +7.50       \\
\cline{2-6}
                           &              & \multicolumn{4}{c}{BEV AP}                         \\
\cline{2-6}
                           & PV-RCNN\cite{PVRCNN}      & 77.45   & 94.64      & 80.39       & 55.39        \\
                           & \textbf{Ours}         & \textbf{82.18}       & \textbf{97.48}          & \textbf{82.51}           & \textbf{64.86}          \\
\cline{2-6}
                           & \textit{Improvement}  & +4.73   & +2.84     & +2.12     & +9.47\\                           
\bottomrule
\end{tabular}
}
% \vspace{-5mm}
\end{table}

\subsection{Results on Waymo Open Dataset}
\textbf{Evaluation Metric.}
The WOD official evaluation toolkit provides two difficulty levels: LEVEL\_1 for boxes with more than five LiDAR points, and LEVEL\_2 for boxes with at least one LiDAR point. The true positive IoU threshold is also set to 0.7.

To show the detection performance in the large-scale point cloud scene that includes more object instances in the full \ang{360} fields, we evaluate our method on both LEVEL\_1 and LEVEL\_2 of the newly released WOD with 3D AP and BEV AP, respectively. Table~\ref{ap_waymo_val} shows that our method consistently outperforms all the other methods with a significant improvement across all the metrics. The detection difficulty generally increases as the object getting far away from the LiDAR sensor due to the fewer points that can be captured. The performance of our method on LEVEL\_1 and LEVEL\_2 improves much more significantly along with the distance. Our method demonstrates a more significant improvement on the large-scale point cloud scenes in WOD. Unlike the sparse point feature extraction in PV-RCNN, the proposed voxel-to-point decoder enables the effective extraction of the fine-grained point features for all raw points in a residual learning manner, which is the key to the performance gains when the model is assembled with the detection refinement as shown in the ablation study. The qualitative results are shown at the bottom of Figure~\ref{qualitative_results}.

\section{Conclusion}
We present an IoU-guided two-stage 3D object detector with a voxel-to-point decoder. We use the voxel-to-point decoder to effectively extract the fine-grained point features, which is crucial to the performance gains in our two-stage detector. The proposed 3D RoI pooling is effective in refining the proposal boxes and estimating the IoUs. We use a simple and efficient method for aligning the estimated IoUs to the refined proposal boxes, thereby further improving the localization accuracy. Experimental results on KITTI and Waymo Open Dataset demonstrate that our method outperforms state-of-the-art methods, and each component in our method is effective with performance gains.

\begin{acks}
This work was supported by the National Key Research and Development Program of China under Grant 2018YFE0183900.
\end{acks}

% \bibliographystyle{ACM-Reference-Format}


% ##############

\end{document}